\documentclass{article}

     \usepackage[nonatbib,final]{ml4eng_2020}

\usepackage[utf8]{inputenc} 
\usepackage[T1]{fontenc}    
\usepackage{hyperref}       
\usepackage{url}            
\usepackage{booktabs}       
\usepackage{amsfonts}       
\usepackage{nicefrac}       
\usepackage{microtype}      
\usepackage{graphicx}
\usepackage{xcolor}
\usepackage{comment}
\usepackage{caption}
\usepackage{subcaption}
\usepackage{amsmath}
\usepackage{tabularx}

\def\sm#1{{\color{black} {#1}}}

\title{Learning to Identify Drilling Defects in Turbine Blades with Single Stage Detectors}

\author{
  Andrea Panizza$^{1}$\thanks{Corresponding author.}, Szymon Tomasz Stefanek$^{2}$, Stefano Melacci$^{2,3}$, \\ \textbf{Giacomo Veneri}$^{1}$, \textbf{Marco Gori}$^{2,3,4}$ \\
  $^{1}$Department of Artificial Intelligence, Baker Hughes,  Florence, Italy\\
  $^{2}$Siena Artificial Intelligence Lab, Siena, Italy\\
  $^{3}$DIISM, University of Siena, Siena, Italy \\
  $^{4}$Maasai, Universit\`{e} C\^{o}te d'Azur, Nice, France \\
  \texttt{andrea.panizza@bakerhughes.com}, \texttt{sts@pragma.software},\\
  \texttt{mela@diism.unisi.it}, \texttt{giacomo.veneri@bakerhughes.com},\\
  \texttt{marco.gori@unisi.it}
}

\begin{document}

\maketitle

\begin{abstract}
  Nondestructive testing (NDT) is widely applied to defect identification of turbine components during manufacturing and operation. Operational efficiency is key for gas turbine OEM (Original Equipment Manufacturers). Automating the inspection process as much as possible, while minimizing the uncertainties involved, is thus crucial. We propose a model based on RetinaNet to identify drilling defects in X-ray images of turbine blades. 
  The application is challenging due to the large image resolutions in which defects are very small and hardly captured by the commonly used anchor sizes, and also due to the small size of the available dataset.
  As a matter of fact, all these issues are pretty common in the application of Deep Learning-based object detection models to industrial defect data. We overcome such issues using open source models, splitting the input images into tiles and scaling them up, applying heavy data augmentation, and optimizing the anchor size and aspect ratios with a differential evolution solver. We validate the model with $3$-fold cross-validation, showing a very high accuracy in identifying images with defects. We also define a set of best practices which can help other practitioners overcome similar challenges.
\end{abstract}

\section{Introduction}

Non-destructive testing (NDT) and structural health monitoring (SHM) techniques are of critical importance for gas turbine OEMS (Original Equipment Manufacturers) \cite{Mevissen2019}. Due to the harsh operating environment, defects in turbine blades can impact the operating life, safety and performance of the gas turbine, so they have to be detected before assembly \cite{Aust_2019}. Internal defects, such as those which may be generated during the drilling of cooling holes, cannot be detected by visual means, and require the use of NDT techniques \cite{Li:2017:1354-2575:364, Chen2019}. One such technique is to acquire X-ray scans of the blades from different viewing angles. The resulting scans have to be analyzed by specialized workers, and the analysis is tedious and time-consuming, especially if large batches of blades are to be examined. Automating the inspection process is thus key to increased operational efficiency. 
In the last years, deep learning has dramatically improved the state-of-the-art in speech recognition, visual object recognition, object detection \cite{liu_deep_2020} and many other domains, such as drug discovery and genomics \cite{lecun_deep_2015}. 

In this work, we apply a single stage detector, RetinaNet \cite{RetinaNet2017} to the problem of detecting drilling defects in turbine blades. The application of deep learning to defect detection in turbine blades has only started recently, and it has been confined to visible light or ultrasonic images of wind turbine blades \cite{shihavuddin_wind_2019,yu_defect_2020,li_investigation_2021}. To the best of our knowledge, this is the first time when deep learning is applied to the analysis of defects in X-ray images of gas turbine blades. This is a significant challenge because of the following issues:

\begin{itemize}
\item The starting images are very large (grayscale images in a proprietary format -- resolution $8496\times6960$).
\item \sm{T}he drilling defects are very small (about $~14\times18$ pixels on average), and they are hard to detect for untrained humans. Detecting small objects is known to be a hard problem even for state-of-the-art object detection models \cite{Ji2020,rabbi_small-object_2020}.
\item The internal dataset we have is fairly small by deep learning standards (694 images, of which 134 with at least one defect). Given the necessity for OEMs to protect their Intellectual Property (IP), there are no public datasets of defects in gas turbine blades with which to integrate our internal one. 
\end{itemize}

We overcome these issues by splitting the initial images into smaller tiles, scaling up the tiles so that the annotations become of comparable or larger size than the smallest anchors, and using random data  augmentation, as well as anchor scale and aspect ratio optimization, in order to maximize the performance of the classifier. With these modifications, we show that our pipeline is capable of detection with almost human-level accuracy. Also, thanks to the usage of a single stage detector, which is faster than two-stage detectors, we keep inference time low, making the tool usable for industrial usage. 

The structure of the paper is as follows: we first describe the dataset (Section~\ref{data}) and the neural models (Section~\ref{Model}), followed by our experimental results (Section~\ref{results}. Finally, conclusions and directions for future research are reported (Section~\ref{sec:concl}).

\section{Dataset}
\label{data}
\begin{figure}[h]
\centering
  \includegraphics[width=0.4\textwidth]{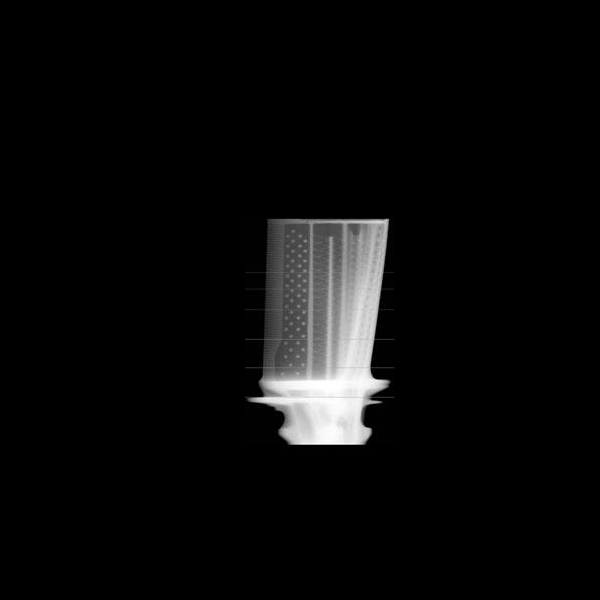}
  \caption{A sample X-ray view of a generic turbine blade. High resolution, large uninformative background region.}
  \label{fig:xray}
\end{figure}

The blades in the dataset are first stage high pressure turbine blades. 
The X-ray images are taken at our 
manufacturing plant, where the drilling process is executed. We have 694 X-ray images, of which 134 with at least one defect. These images correspond to 498 distinct blades, of which 88 with at least one defect. This means that for a few blades multiple X-ray images from different viewing angles were acquired. A X-ray image of a generic turbine blade (not one of ours) is shown in Fig.\ref{fig:xray}.

\begin{figure}
    \centering
    \begin{subfigure}{0.3\textwidth}
        \includegraphics[width=\textwidth]{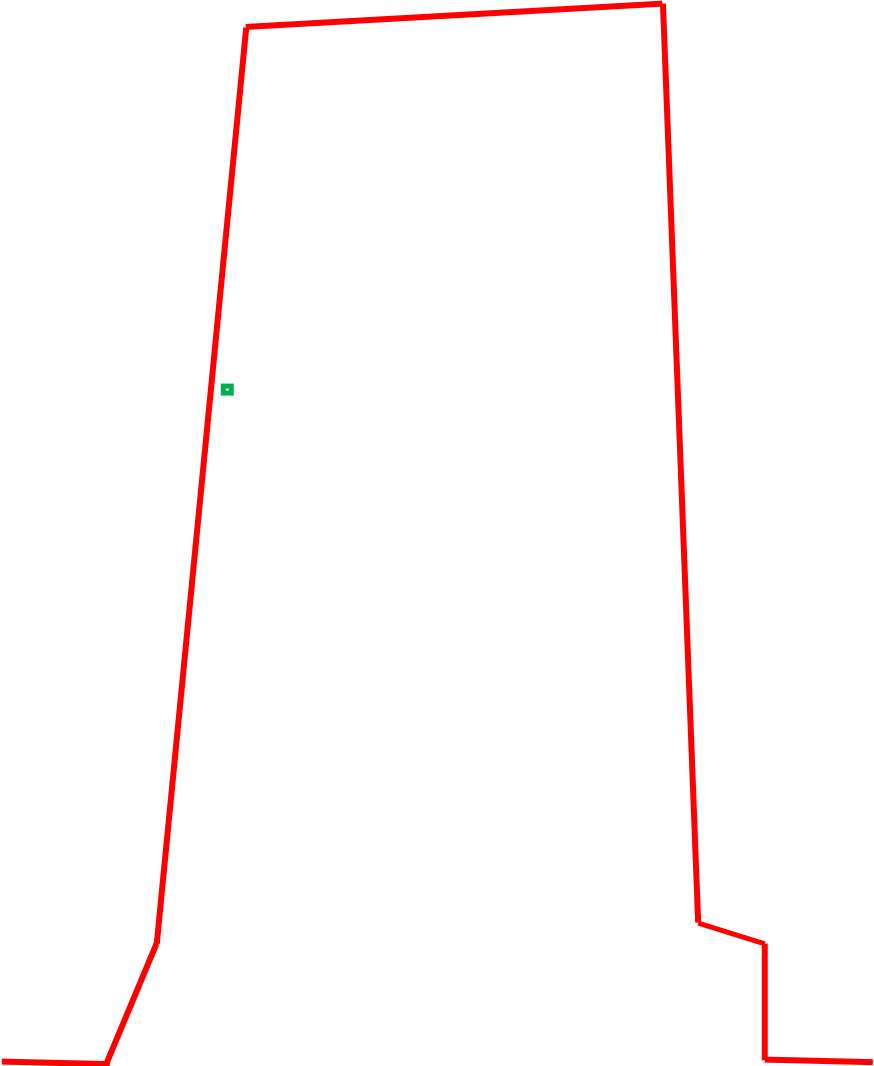}
    \end{subfigure}
    \hskip 7mm
    \begin{subfigure}{0.45\textwidth}
        \includegraphics[width=\textwidth]{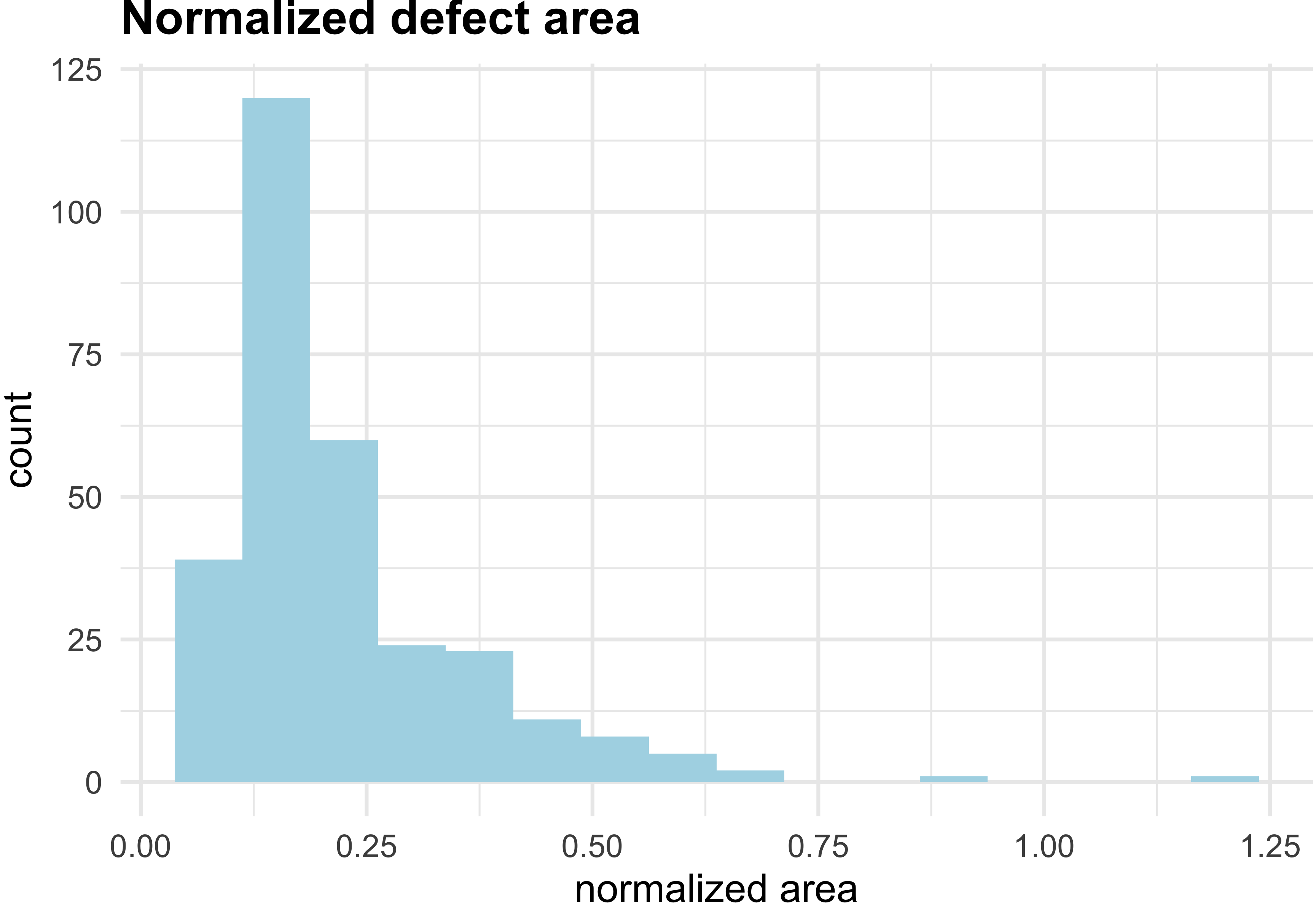}
    \end{subfigure}
        \caption{\sm{L}eft: relative size of a defect with respect to the aerofoil. \sm{The contour of the airfoil is reported in red, while the defect area is highlighted in green.} Right: histogram of defect areas normalized by smallest anchor area.}
    \label{fig:defect}
\end{figure}

The high-quality X-ray scans are very large ($8496\times6960$ pixels), 
and they include a large uninformative background region that has no relevance in the task of defect detection.
The scans highlight holes that are drilled only in the part of the blade which is exposed to the gas flow (aerofoil), but not in the lower part which is secured in the rim of the wheel (foot). 
 We used the \texttt{threshold} and \texttt{findContours} methods of the OpenCV library \cite{opencv_library} to automatically crop out the aerofoil, which results in a $1900\times1500$ image. The defects are extremely small with respect to the aerofoil, as it can be seen in the left part of Fig.~\ref{fig:defect}. Each defect has been accurately annotated by highly experienced manufacturing engineers, for a total of 294 annotations. The right part of Fig.~\ref{fig:defect} shows the histogram of the defect areas normalized with respect to the area of the smallest RetinaNet anchor, which is $32\times32$ pixels. Clearly, most of the defects are smaller than the smallest anchor, which makes 
 their detection extremely complicated.

We postpone to Section~\ref{Model} the description of the RetinaNet model, and we only describe the dataset we used to validate this intuition. We select\sm{ed} all the 134 positive (with at least one defect) images,  while for the negative (no defect) images, we select\sm{ed} only a random subsample of 148 images, for a total dataset size $N=282$. The reason why we discard\sm{ed} most negative images is that the accuracy of one-stage detectors is affected by the problem of extreme foreground-background class imbalance. RetinaNet implements a focal loss to address this issue \cite{RetinaNet2017}. However, even with focal loss, negative (no defect at all) images remain problematic, thus we \sm{kept} a ratio $\frac{\text{nr. of negative images}}{\text{nr. of positive images}}=1.1$. We \sm{divided} the dataset in\sm{to} a training set with $196$ images and a validation set with $86$ images. After 120 epochs with a minibatch size of 2 (limited by the large size of the images) on a NVIDIA Tesla V100 32Gb GPU, the mAP on the validation set, as computed by the RetinaNet implementation we used, was 0.06, i.e., the model did not learn anything useful. 

\begin{figure}
    \centering
    \includegraphics[width=0.4\textwidth]{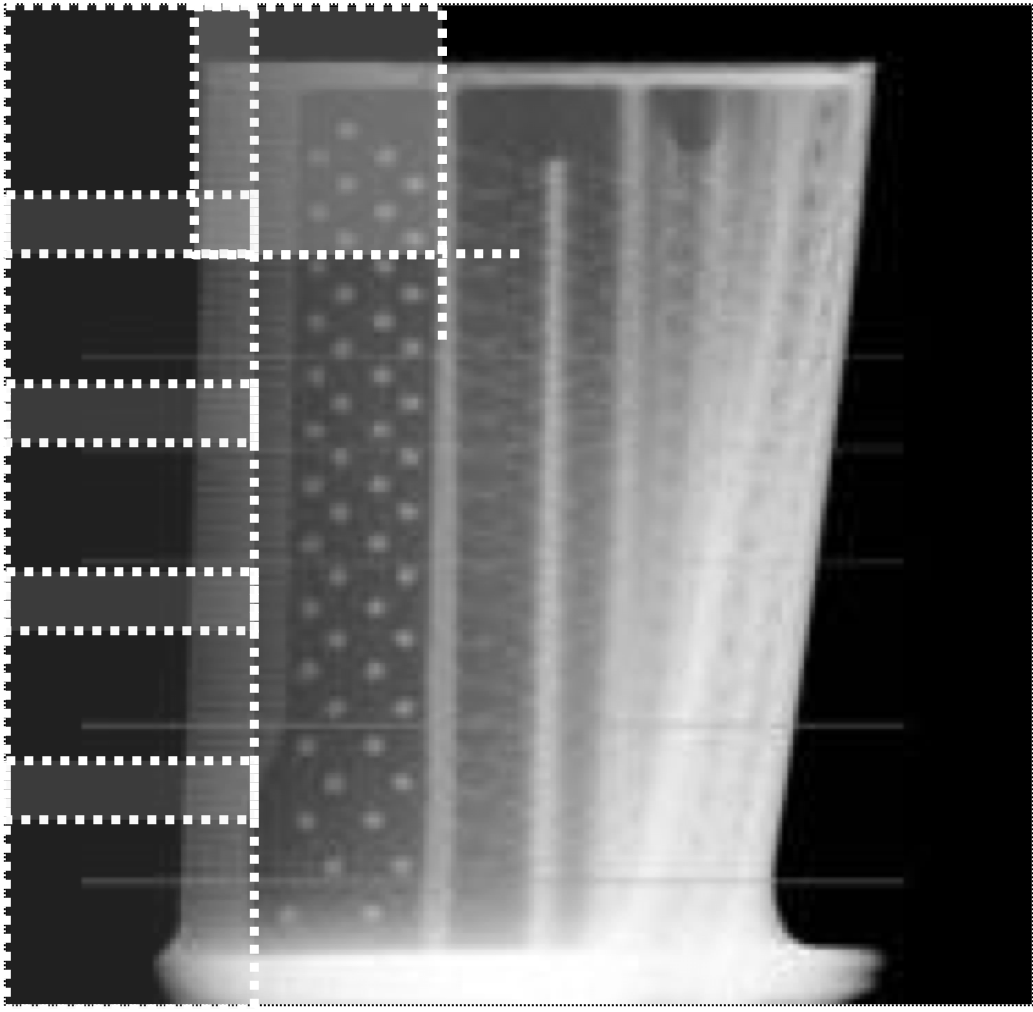}
    \hskip 5mm
    \includegraphics[width=0.5\textwidth]{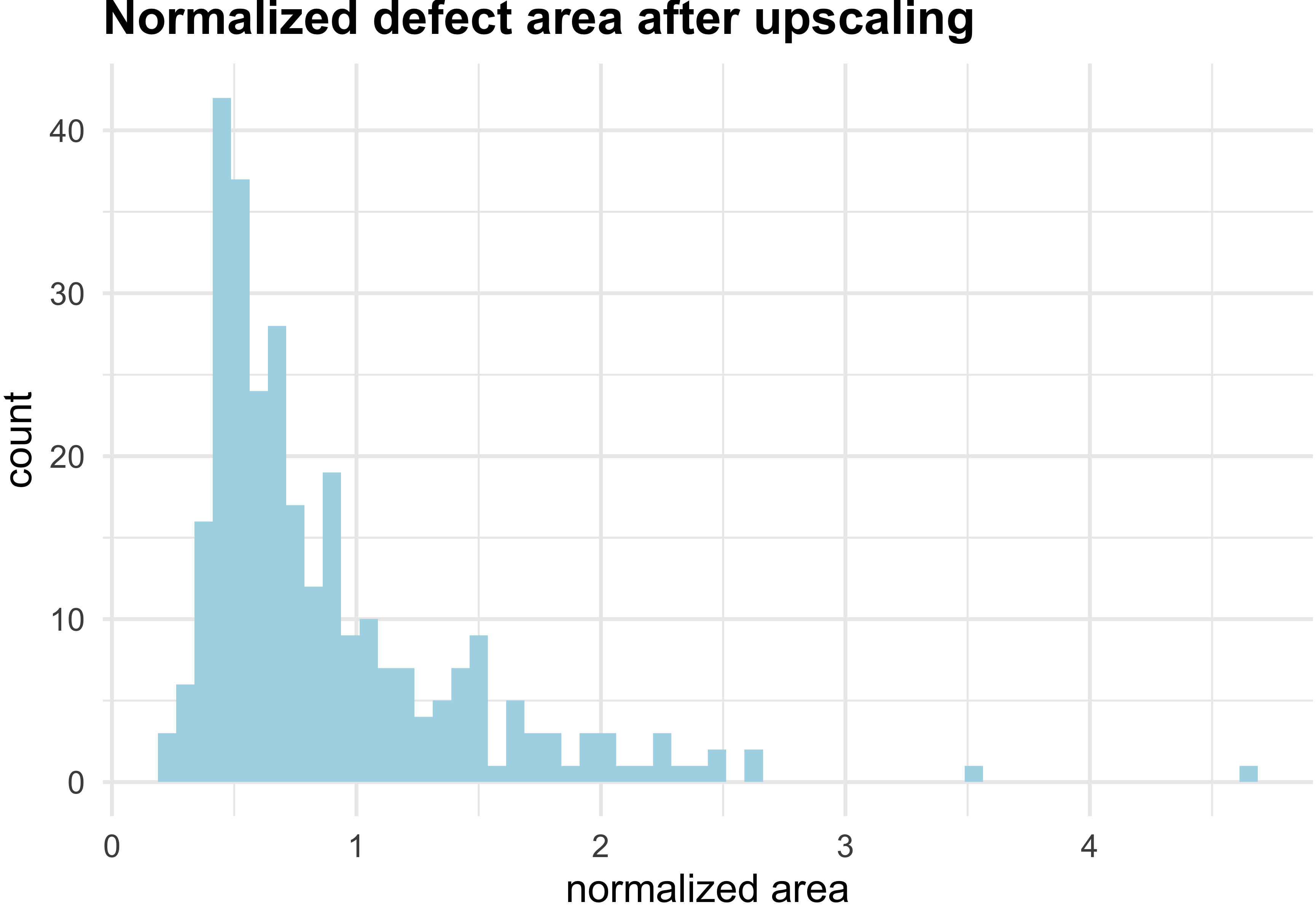}    
    \caption{Left: subdivision of an image in overlapping tiles \sm{(we show the boundaries of some tiles using dotted lines)}. Right: histogram of defect areas normalized by the smallest anchor area, after scaling up height and width of each tile by $2\times$.}
    \label{fig:tiling}
\end{figure}

To overcome this issue, we \sm{divided} each image in\sm{to} $5\times5$ overlapping tiles of $500\times600$ pixels, as shown schematically in Fig.~\ref{fig:tiling} (left), with an horizontal overlap of $250$ pixels and a vertical one of $325$. The overlap is important because object detection models are often less sensitive towards the boundaries of an image. Thanks to the overlap, if a defect is near the boundary of a tile, it \sm{is} well inside the interior of at least one other tile, thus the model should be able to detect it. The only exception, of course, is \sm{the case in which} the defect is near the boundary of the original (unsplitted) image. Clearly, since the number of drilling defects in the positive images is limited, most of the tiles end up being empty. Thus, we balance\sm{d} again the number of positive and negative tiles, ending up with the statistics \sm{reported} in Table~\ref{tab:tiled_dataset}. Note that to avoid any data leakage, all the retained tiles from a given image 
\sm{either belong to to the training or the validation set.}
At training time, we scale\sm{d} the images and annotations up by $2\times$ in height and width, so that the new histogram of defect areas normalized by \sm{the} smallest anchor area become as shown in Fig.~\ref{fig:tiling} (right). In Section~\ref{results} we 
\sm{will evaluate the effects of each of these operations}.
\begin{table}
  \caption{Dataset structure.}
  \label{tab:tiled_dataset}
  \vspace{\baselineskip}
  \centering
  \begin{tabular}{llll}
    \toprule
    Set & Positive & Negative & Total \\
    \midrule
    Train & 363 & 400  & 763 \\
    \midrule
    Validation  & 153 & 175  & 328  \\
    \bottomrule
  \end{tabular}
\end{table}

\section{Model}
\label{Model}
RetinaNet \cite{RetinaNet2017} \sm{is} a single stage model for object detection
\sm{that}
uses a focal loss to address the common problem of class imbalance in detection tasks. RetinaNet \sm{exploits} a feature pyramid network with a top-down architecture to detect objects at different scales \cite{FPN2017}. The implementation we use\sm{d} is a fork of the  Fizyr\footnote{https://github.com/fizyr/keras-retinanet} implementation.
\sm{The} RetinaNet model architecture \sm{consists of} four major components: 
\begin{itemize}
\item a \emph{bottom-up pathway}, the backbone network which calculates the feature maps at different scales. We use\sm{d} a ResNet-50 model \cite{ResNets2016}.
\item \emph{top-down pathway and lateral connections}, the top down pathway upsamples the spatially coarser feature maps from higher pyramid levels, and the lateral connections merge the top-down layers and the bottom-up layers with the same spatial size.
\item a \emph{classification subnetwork} predicts the probability of an object being present at each spatial location for each anchor box and object class.
\item a \emph{regression subnetwork} regresses the offset for the bounding boxes from the anchor boxes to a nearby ground-truth object, if one exists.
\end{itemize}
For more details, we point to the original RetinaNet paper. Regarding our specific implementation, the model is pretrained on the MS COCO dataset\cite{COCO_2014}. We use\sm{d} random data augmentation to improve the model performance. We trained the model for 40 epochs.

\subsection{Anchor Optimization}
\label{anchor_opt}
The anchor configuration is \sm{a crucial feature in the detection process}, and we find the default anchor sizes (32, 64, 128, 256 and 512), aspect ratios (0.5, 1, 2) and scales (1, 1.2, 1.6) to be ineffective for detecting our small defects. We employ\sm{ed} the differential evolution search algorithm used by \cite{zlocha_improving_2019} to optimize ratios and scales of anchors on the validation set. The algorithm iteratively improves a population of candidate solutions with regard to an objective function. The implementation is freely available on GitHub\footnote{https://github.com/martinzlocha/anchor-optimization/}.

\section{Results}
\label{results}

We present the results of the compared settings in Table \ref{tab:results}. The model trained on the original images, without tiling and upscaling, delivers useless results. The model retrained on the splitted images shows a minimal improvement, with a mAP on the validation set of 0.10, since the vast majority of the annotations is smaller than the smallest default anchor. Adding $2\times$ upscaling drives the mAP up to 0.73. This is much better, but still not enough to deploy the model in production. Finally, including anchor optimization brings the mAP to 0.90, which meets our requirements. The results are summarized in Table~\ref{tab:results} (mAP).

We also show the results of a different metric, which is more closely related to our business goals (Table~\ref{tab:results}, Accuracy). We care more about catching blades with one or more defects, rather than localizing the defects exactly. 
For this reason, we use the following definition of accuracy, for which an image is correctly classified if the following conditions are both true:
\begin{itemize}
\item The model predicts a number of defects that is at least half of the number of annotations for the considered image.
\item Let $S$ be the union of the predicted bounding boxes, and $G$ the union of the annotations. The Intersection over Union (IoU) of $G$ and $S$ must be more than 0.2.
\end{itemize}
To estimate the generalization error, we computed this metric in a $3$-fold cross-validation setting. The resulting accuracy averaged over the 3 folds is \textbf{0.94} (with negligible standard deviation) -- very close to human accuracy, and large enough for our goals.

\begin{table}
  \caption{Performance metrics.}
  \label{tab:results}
  \vspace{\baselineskip}
  \centering
  \begin{tabularx}{\textwidth}{Xll}
    \toprule
    Setting & mAP & Accuracy \\
    \midrule
    Original images $(1500\times1900)$ & 0.06 & 0.56 \\
    \midrule
    Splitted tiles $(500\times600)$ & 0.10 & 0.53 \\
    \midrule
    Splitted tiles + 2x upscaling $(1000\times1200)$ & 0.73 & 0.90  \\
    \midrule
    Splitted tiles + 2x upscaling $(1000\times1200)$ 
    + anchor optimization & 0.90 & 0.94 \\
    \bottomrule
  \end{tabularx}
\end{table}

\section{Conclusion}
\label{sec:concl}
We developed a model to identify drilling defects in X-ray images of gas turbine blades. Simply training a very powerful object detection model such as RetinaNet on our dataset did not deliver any usable results. We identified the key issues, which were the large size of the images, preventing the usage of large minibatches, the small average size of the annotations, and the use of default anchor scales and aspect ratios, which were not optimized for our data distribution. The final model delivers very promising results. Before deploying the model in production, we will collect more data to better estimate the generalization error of the model. This work also opens the way to a more general application of deep learning methods to defect detection in gas turbine components.

\bibliographystyle{IEEEtran}
\bibliography{blade-x}

\end{document}